\documentclass[runningheads]{llncs}
\usepackage{graphicx}

\usepackage{hyperref}
\hypersetup{breaklinks=true,colorlinks,citecolor=black}

\usepackage{tikz}
\usepackage{comment}
\usepackage{amsmath,amssymb} 

\usepackage{colortbl}
\usepackage{enumitem}
\usepackage{bm}
\usepackage{url}
\usepackage{bbm}
\usepackage{multirow}
\usepackage{color}
\usepackage{tabu,comment}
\usepackage{array, eucal}
\usepackage[ruled,vlined]{algorithm2e}
\usepackage{algorithmic}
\usepackage{orcidlink}
\usepackage{tabularx,multirow,makecell,setspace,color,booktabs}
\usepackage{array}
\usepackage[utf8]{inputenc}
\usepackage{color}
\usepackage{caption}
\usepackage[american]{babel}
\usepackage{subcaption}
\usepackage{babel}[English]

\usepackage[accsupp]{axessibility}  

\graphicspath{{./Figures/}}

\makeatletter
\renewcommand\paragraph{\@startsection{paragraph}{4}{\z@}%
                       {-12\p@ \@plus -4\p@ \@minus -4\p@}%
                       {-0.5em \@plus -0.22em \@minus -0.1em}%
                       {\normalfont\normalsize}}
\makeatother

\DeclareMathOperator{\softmax}{softmax}

\begin{document}
\pagestyle{headings}
\mainmatter
\def\ECCVSubNumber{829}  

\title{FADE: Fusing the Assets of Decoder and Encoder for Task-Agnostic Upsampling} 

\titlerunning{FADE: Towards Task-Agnostic Feature Upsampling}
%
\author{
Hao Lu\orcidlink{0000-0003-3854-8664} \and
Wenze Liu\orcidlink{0000-0002-1510-6196} \and
Hongtao Fu\orcidlink{0000-0002-6692-0913} \and
Zhiguo Cao\thanks{Corresponding author}\orcidlink{0000-0002-9223-1863}
}
\authorrunning{H. Lu et al.}
%
\institute{School of Artificial Intelligence and Automation, \\
Huazhong University of Science and Technology, Wuhan 430074, China 
\email{\{hlu,wzliu,htfu,zgcao\}@hust.edu.cn}}
\maketitle

\begin{abstract}

We consider the problem of task-agnostic feature upsampling in dense prediction where an upsampling operator is required to facilitate both region-sensitive tasks like semantic segmentation and detail-sensitive tasks such as image matting. Existing upsampling operators often can work well in either type of the tasks, but not both. In this work, we present FADE, a novel, plug-and-play, and task-agnostic upsampling operator. FADE benefits from three design choices:
i) considering encoder and decoder features jointly in upsampling kernel generation;  
ii) an efficient semi-shift convolutional operator that enables granular control over how each feature point contributes to upsampling kernels; 
iii) a decoder-dependent gating mechanism for enhanced detail delineation.
We first study the upsampling properties of FADE on toy data and then evaluate it on large-scale semantic segmentation and image matting.
In particular, FADE reveals its effectiveness and task-agnostic characteristic by consistently outperforming recent dynamic upsampling operators in different tasks. It also generalizes well across convolutional and transformer architectures with little computational overhead. Our work additionally provides thoughtful insights on what makes for task-agnostic upsampling. Code is available at: \url{http://lnkiy.in/fade_in}




\end{abstract}

\section{Introduction}

\begin{figure}[t]
	\centering
	\includegraphics[width=\linewidth]{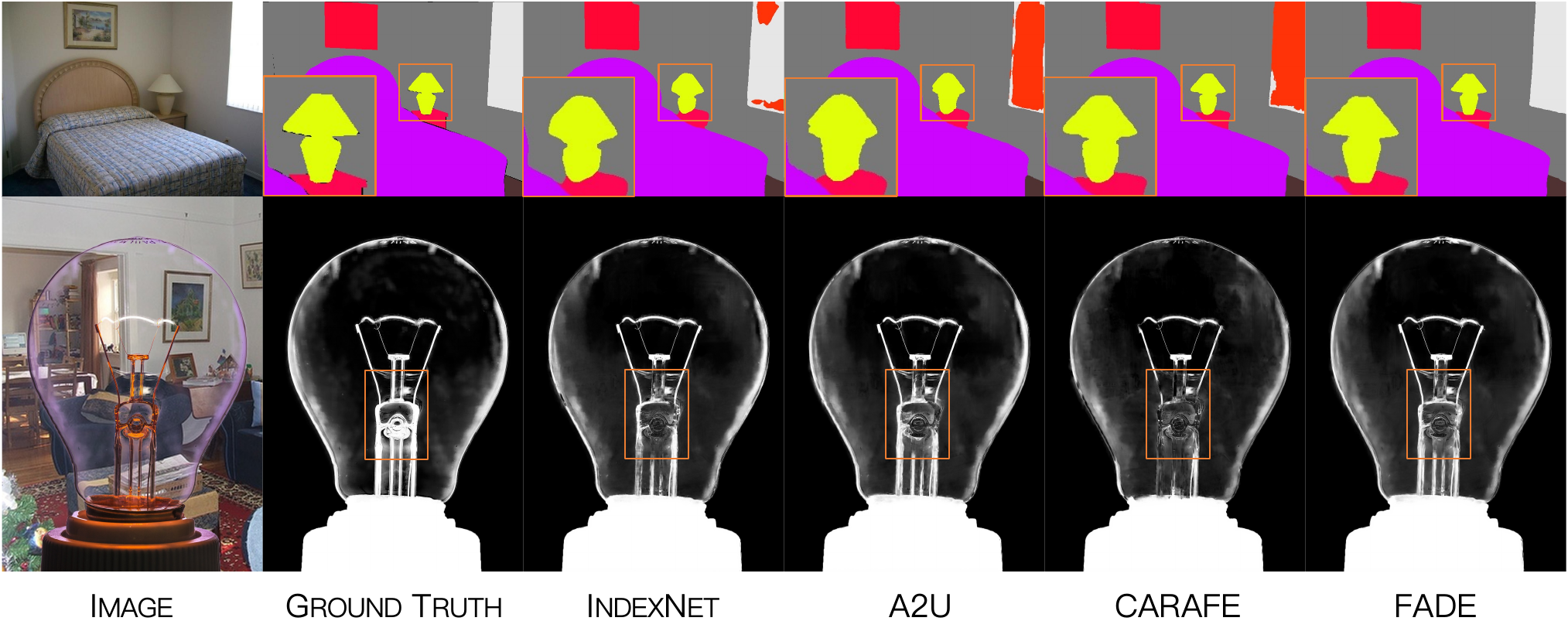}
	\caption{\textbf{Inferred segmentation masks and alpha mattes with different upsampling operators}. The compared operators include IndexNet~\cite{lu2019indices}, A2U~\cite{dai2021learning}, CARAFE~\cite{jiaqi2019carafe}, and our proposed FADE. Among all competitors, only FADE generates both the high-quality mask and the alpha matte.}
	\label{fig:seg_matte_intro}
\end{figure}

Feature upsampling, which aims to recover the spatial resolution of features, is an indispensable stage in many dense prediction models~\cite{ronneberger2015u,badrinarayanan2017segnet,xiao2018unified,wang2020deep,zheng2021rethinking,xie2021segformer}. Conventional upsampling operators, such as nearest neighbor (NN) or bilinear interpolation~\cite{lin2017refine}, deconvolution~\cite{zeiler2014visualizing}, and pixel shuffle~\cite{shi2016real}, often have a preference of a specific task. For instance, bilinear interpolation is favored in semantic segmentation~\cite{chen18v3,xie2021segformer}, and pixel shuffle is preferred in image super-resolution~\cite{Ignatov_2021_CVPR}. 

A main reason is that each dense prediction task has its own focus: some tasks like semantic segmentation~\cite{long2015fully} and instance segmentation~\cite{he2017mask} are region-sensitive, while some tasks such as image super-resolution~\cite{dong2015image} and image matting~\cite{xu2017deep,lu2019indices} are detail-sensitive. If one expects an upsampling operator to generate semantically consistent features such that a region can share the same class label, it is often difficult for the same operator to recover boundary details simultaneously, and vice versa. Indeed empirical evidence shows that bilinear interpolation and max unpooling~\cite{badrinarayanan2017segnet} have inverse behaviors in segmentation and matting~\cite{lu2019indices,lu2022index}, respectively.

In an effort to evade `trials-and-errors' from choosing an upsampling operator for a certain task at hand, there has been a growing interest in developing a generic upsampling operator for dense prediction recently~\cite{mazzini2018guided,tian2019decoders,jiaqi2019carafe,wang2020carafe++,lu2019indices,lu2022index,dai2021learning}. For example, CARAFE~\cite{jiaqi2019carafe} demonstrates its benefits on four dense prediction tasks, including object detection, instance segmentation, semantic segmentation, and image inpainting. IndexNet~\cite{lu2019indices} also boosts performance on several tasks such as image matting, image denoising, depth prediction, and image reconstruction. However, a comparison between CARAFE and IndexNet~\cite{lu2022index} indicates that neither CARAFE nor IndexNet can defeat its opponent on both region- and detail-sensitive tasks (CARAFE outperforms IndexNet on segmentation, while IndexNet is superior than CARAFE on matting), which can also be observed from the inferred segmentation masks and alpha mattes in Fig.~\ref{fig:seg_matte_intro}. This raises an interesting question: \textit{Does there exist a unified form of upsampling operator that is truly task-agnostic?}


To answer the question above, we present FADE, a novel, plug-and-play, and task-agnostic upsampling operator which Fuses the Assets of Decoder and Encoder (FADE). The name also implies its working mechanism: upsampling features in a `fade-in' manner, from recovering spatial structure to delineating subtle details. In particular, we argue that an ideal upsampling operator should be able to preserve the semantic information and compensate the detailed information lost due to downsampling. The former is embedded in decoder features; the latter is abundant in encoder features. Therefore, we hypothesize that it is the insufficient use of encoder and decoder features bringing the task dependency of upsampling, and our idea is to design FADE to make the best use of encoder and decoder features, inspiring the following insights and contributions:
\begin{enumerate}

    \item[i)] By exploring why CARAFE works well on region-sensitive tasks but poorly on detail-sensitive tasks, and why IndexNet and A2U~\cite{dai2021learning} behave conversely, we observe that what features (encoder or decoder) to use to generate the upsampling kernels matters. Using decoder features can strengthen the regional continuity, while using encoder features helps recover details. It is thus natural to seek whether combining encoder and decoder features enjoys both merits, which underpins the core idea of FADE.
    
    \item[ii)] To integrate encoder and decoder features, a subsequent problem is how to deal with the resolution mismatch between them. A standard way is to implement UNet-style fusion~\cite{ronneberger2015u}, including feature interpolation, feature concatenation, and convolution. However, we show that this naive implementation can have a negative effect on upsampling kernels. 
    To solve this, we introduce a semi-shift convolutional operator that unifies channel compression, concatenation, and kernel generation. Particularly, it allows granular control over how each feature point participates in the computation of upsampling kernels. The operator is also fast and memory-efficient due to direct execution of cross-resolution convolution, without explicit feature interpolation for resolution matching. 
    
    \item[iii)] To enhance detail delineation, we further devise a gating mechanism, conditioned on decoder features. The gate allows selective pass of fine details in the encoder features as a refinement of upsampled features.
    
\end{enumerate}

We conduct experiments on five data sets covering three dense prediction tasks. We first validate our motivation and the rationale of our design through several toy-level and small-scale experiments, such as binary image segmentation on Weizmann Horse~\cite{borenstein2002class}, image reconstruction on Fashion-MNIST~\cite{xiao2017fashion}, and semantic segmentation on SUN RGBD~\cite{song2015sun}. We then present a thorough evaluation of FADE on large-scale semantic segmentation on ADE20K~\cite{zhou2017scene} and image matting on Adobe Composition-1K~\cite{xu2017deep}. FADE reveals its task-agnostic characteristic by consistently outperforming state-of-the-art upsampling operators on both region- and detail-sensitive tasks, while also retaining the lightweight property by appending relatively few parameters and FLOPs. It has also good generalization across convolutional and transformer architectures~\cite{he2016deep,xie2021segformer}.

To our knowledge, FADE is the first task-agnostic upsampling operator that performs favorably on both region- and detail-sensitive tasks. 

\section{Related Work}

\paragraph{\textbf{Feature Upsampling.}} Unlike joint image upsampling~\cite{tomasi1998bilateral,he2010guided}, feature upsampling operators are mostly developed in the deep learning era, to respond to the need for recovering spatial resolution of encoder features (decoding). Conventional upsampling operators typically use fixed/hand-crafted kernels. For instance, the kernels in the widely used NN and bilinear interpolation are defined by the relative distance between pixels. Deconvolution~\cite{zeiler2014visualizing}, \textit{a.k.a.}\ transposed convolution, also applies a fixed kernel during inference, despite the kernel parameters are learned. Pixel shuffle~\cite{shi2016real} instead only includes memory operations but still follows a specific rule in upsampling by reshaping the depth channel into the spatial channels. Among hand-crafted operators, unpooling~\cite{badrinarayanan2017segnet} perhaps is the only operator that has a dynamic upsampling behavior, \textit{i.e.}, each upsampled position is data-dependent conditioned on the $\max$ operator. Recently the importance of the dynamic property has been proved by some dynamic upsampling operators~\cite{jiaqi2019carafe,lu2019indices,dai2021learning}. CARAFE~\cite{jiaqi2019carafe} implements context-aware reassembly of features, IndexNet~\cite{lu2019indices} provides an indexing perspective of upsampling, and A2U~\cite{dai2021learning} introduces affinity-aware upsampling. At the core of these operators is the data-dependent upsampling kernels whose kernel parameters are predicted by a sub-network. This points out a promising direction from considering generic feature upsampling. FADE follows the vein of dynamic feature upsampling.

\paragraph{\textbf{Dense Prediction.}} Dense prediction covers a broad class of per-pixel labeling tasks, ranging from mainstream object detection~\cite{ren2015faster}, semantic segmentation~\cite{long2015fully}, instance segmentation~\cite{he2017mask}, and depth estimation~\cite{eigen2014depth} to low-level image restoration~\cite{mao2016image}, image matting~\cite{xu2017deep}, edge detection~\cite{xie2015holistically}, and optical flow estimation~\cite{teed2020raft}, to name a few. An interesting property about dense prediction is that a task can be region-sensitive or detail-sensitive. The sensitivity is closely related to what metric is used to assess the task. In this sense, semantic/instance segmentation is region-sensitive, because the standard Mask Intersection-over-Union (IoU) metric~\cite{everingham2010pascal} is mostly affected by regional mask prediction quality, instead of boundary quality. On the contrary, image matting can be considered detail-sensitive, because the error metrics~\cite{rhemann2009perceptually} are mainly computed from trimap regions that are full of subtle details or transparency. Note that, when we emphasize region sensitivity, we do not mean that details are not important, and vice versa. In fact, the emergence of Boundary IoU~\cite{cheng2021boundary} implies that the limitation of a certain evaluation metric has been noticed by our community. The goal of developing a task-agnostic upsampling operator capable of both regional preservation and detail delineation can have a board impact on a number of dense prediction tasks. In this work, we mainly evaluate upsampling operators on semantic segmentation and image matting, which may be the most representative region- and detail-sensitive task, respectively.

\section{Task-Agnostic Upsampling: A Trade-off Between Semantic Preservation and Detail Delineation}
\label{sec:insights}

Before we present FADE, we share some of our view points towards task-agnostic upsampling, which may be helpful to understand our designs in FADE.

\paragraph{\textbf{How Encoder and Decoder Features Affect Upsampling.}}

In dense prediction models, downsampling stages are involved to acquire a large receptive field, bringing the need of peer-to-peer upsampling stages to recover the spatial resolution, which together constitutes the basic encoder-decoder architecture. During downsampling, details of high-resolution features are impaired or even lost, but the resulting low-resolution encoder features often have good semantic meanings that can pass to decoder features.
Hence, we believe an ideal upsampling operator should appropriately resolve two issues: 1) preserve the semantic information already extracted; 2) compensate as many lost details as possible without deteriorating the semantic information. NN or bilinear interpolation only meets the former. This conforms to our intuition that interpolation often smooths features. A reason is that low-resolution decoder features have no prior knowledge about missing details. Other operators that directly upsample decoder features, such as deconvolution and pixel shuffle, can have the same problem with poor detail compensation. Compensating details requires high-resolution encoder features. This is why unpooling that stores indices before downsampling has good boundary delineation~\cite{lu2019indices}, but it hurts the semantic information due to zero-filling.


\begin{figure}[!t]
	\centering
	\includegraphics[width=\linewidth]{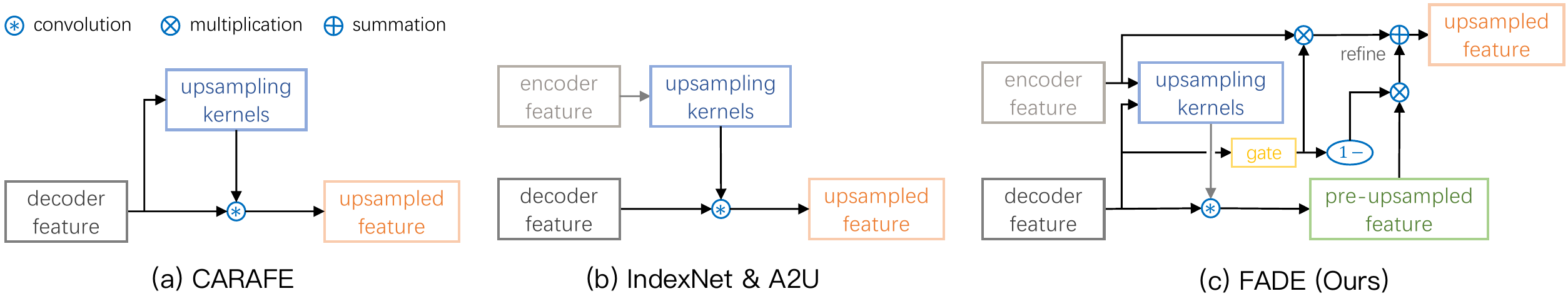}
	\caption{\textbf{Use of encoder and/or decoder features in different upsampling operators}. (a) CARAFE generates upsampling kernels conditioned on decoder features, while (b) IndexNet and A2U generate kernels using encoder features only. By contrast, (c) FADE considers both encoder and decoder features not only in upsampling kernel generation but also in gated feature refinement.}
	\label{fig:concept_map}
\end{figure}

Dynamic upsampling operators, including CARAFE~\cite{jiaqi2019carafe}, IndexNet~\cite{lu2019indices}, and A2U~\cite{dai2021learning}, alleviate the problems above with data-dependent upsampling kernels. Their upsampling modes are illustrated in Fig.~\ref{fig:concept_map}(a)-(b). From Fig.~\ref{fig:concept_map}, it can be observed that, CARAFE generates upsampling kernels conditioned on decoder features, while IndexNet~\cite{lu2019indices} and A2U~\cite{dai2021learning} generate kernels via encoder features. This may explain the inverse behavior between CARAFE and IndexNet/A2U on region- or detail-sensitive tasks~\cite{lu2022index}. In this work, we find that generating upsampling kernels using either encoder or decoder features can lead to suboptimal results, and it is critical \textit{to leverage both encoder and decoder features for task-agnostic upsampling}, as implemented in FADE (Fig.~\ref{fig:concept_map}(c)).


\paragraph{\textbf{How Each Feature Point Contributes to Upsampling Matters.}}

After deciding what the features to use, the follow-up question is how to use the features effectively and efficiently. The main obstacle is the mismatched resolution between encoder and decoder feature maps. One may consider simple interpolation for resolution matching, but we find that this leads to sub-optimal upsampling. Considering the case of applying $\times2$ NN interpolation to decoder features, if we apply $3\times3$ convolution to generate the upsampling kernel, the effective receptive field of the kernel can be reduced to be $<50\%$: before interpolation there are $9$ valid points in a $3\times3$ window, but only $4$ valid points are left after interpolation, as shown in Fig.~\ref{fig:semi-shift_conv}(a). Besides this, there is another more important issue. 
Still in $\times2$ upsampling, as shown in Fig.~\ref{fig:semi-shift_conv}(a), the four windows which control the variance of upsampling kernels w.r.t.\ the $2\times2$ neighbors of high resolution are influenced by the hand-crafted interpolation. 
Controlling a high-resolution upsampling kernel map, however, is blind with the low-resolution decoder feature. It contributes little to an informative upsampling kernel, especially to the variance of the four neighbors in the upsampling kernel map. Interpolation as a bias of that variance can even worsen the kernel generation.
A more reasonable choice may be 
to \textit{let encoder and decoder features cooperate to control the overall upsampling kernel, but let the encoder feature alone control the variance of the four neighbors}. This insight exactly motivates the design of semi-shift convolution (Section \ref{sec:fade}).

\paragraph{\textbf{Exploiting Encoder Features for Further Detail Refinement.}}

Besides helping structural recovery via upsampling kernels, there remains much useful information in the encoder features. Since encoder features only go through a few layers of a network, they preserve `fine details' of high resolution. In fact, nearly all dense prediction tasks require fine details, \textit{e.g.}, despite regional prediction dominates in instance segmentation, accurate boundary prediction can also significantly boost performance~\cite{tang2021look}, not to mention the stronger request of fine details in detail-sensitive tasks. \textit{The demands of fine details in dense prediction need further exploitation of encoder features.} Instead of simply skipping the encoder features, we introduce a gating mechanism that leverages decoder features to guide where the encoder features can pass through.

\section{Fusing the Assets of Decoder and Encoder}
\label{sec:fade}

\paragraph{\textbf{Dynamic Upsampling Revisited.}} Here we review some basic operations in recent dynamic upsampling operators such as CARAFE~\cite{jiaqi2019carafe}, IndexNet~\cite{lu2019indices}, and A2U~\cite{dai2021learning}. Fig.~\ref{fig:concept_map} briefly summarizes their upsampling modes. They share an identical pipeline, \textit{i.e.}, first generating data-dependent upsampling kernels, and then reassembling the decoder features using the kernels. Typical dynamic upsampling kernels are content-aware, but channel-shared, which means each position has a unique upsampling kernel in the spatial dimension, but the same ones are shared in the channel dimension.


CARAFE learns upsampling kernels directly from decoder features and then reassembles them to high resolution. In particular, the decoder features pass through two consecutive convolutional layers to generate the upsampling kernels, of which the former is a channel compressor implemented by $1\times1$ convolution to reduce the computational complexity and the latter is a content encoder with $3\times3$ convolution, and finally the $\softmax$ function is used to normalize the kernel weights. IndexNet and A2U, however, 
adopt more sophisticated modules to leverage the merit of encoder features. Further details can be referred to~\cite{jiaqi2019carafe,lu2019indices,dai2021learning}.

\begin{figure}[!t]
	\centering
	\includegraphics[width=\linewidth]{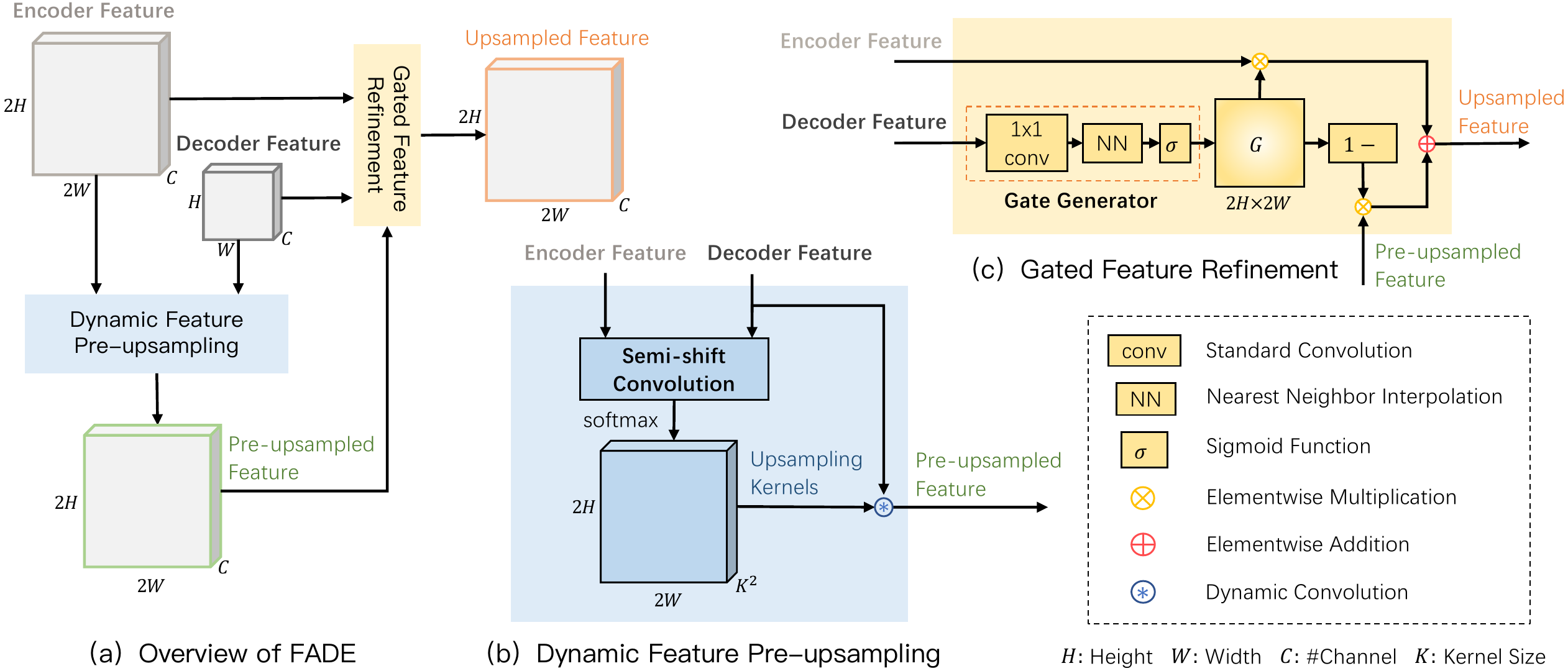}
	\caption{\textbf{Technical pipeline of FADE}. From (a) the overview of FADE, feature upsampling is executed by jointly exploiting the encoder and decoder feature with two key modules. In (b) dynamic feature pre-upsampling, they are used to generate upsampling kernels using a semi-shift convolutional operator (Fig.~\ref{fig:semi-shift_conv}). The kernels are then used to reassemble the decoder feature into pre-upsampled feature. In (c) gated feature refinement, the encoder and pre-upsampled features are modulated by a decoder-dependent gating mechanism to enhance detail delineation before generating the final upsampled feature.}
	\label{fig:pipeline}
\end{figure}

FADE is designed to maintain the simplicity of dynamic upsampling. 
Hence, it generally follows the pipeline of CARAFE, but further optimizes the process of kernel generation with semi-shift convolution, and the channel compressor will also function as a way of pre-fusing encoder and decoder features. In addition, FADE also includes a gating mechanism for detail refinement. The overall pipeline of FADE is summarized in Fig.~\ref{fig:pipeline}.


\paragraph{\textbf{Generating Upsampling Kernels from Encoder and Decoder Features.}}

\begin{figure}[!t]
	\centering
	\includegraphics[width=\linewidth]{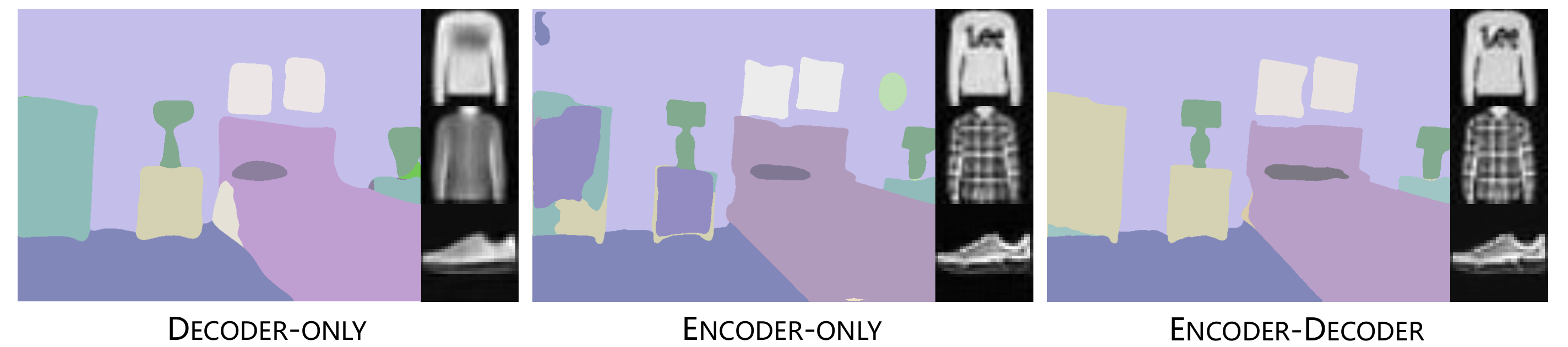}
	\caption{\textbf{Visualizations of inferred mask and reconstructed results on SUN RGBD and Fashion-MNIST}. The decoder-only model generates good regional prediction but poor boundaries/textures, while the encoder-only one is on the contrary. When fusing encoder and decoder features, both region and detail predictions are improved, \textit{e.g.}, the table lamp and stripes on clothes.}
	\label{fig:sunrgbd_visual}
\end{figure}

We first showcase a few visualizations on some small-scale or toy-level data sets to highlight the importance of both encoder and decoder features for task-agnostic upsampling. We choose semantic segmentation on SUN RGBD~\cite{song2015sun} as the region-sensitive task and image reconstruction on Fashion MNIST~\cite{xiao2017fashion} as the detail-sensitive one. We follow the network architectures and the experimental settings in~\cite{lu2022index}. Since we focus on upsampling, all downsampling stages use max pooling. Specifically, to show the impact of encoder and decoder features, in the segmentation experiments, we all use CARAFE but only modify the source of features used for generating upsampling kernels. We build three baselines: 
\mbox{1) \textit{decoder-only}}, the implementation of CARAFE; 
\mbox{2) \textit{encoder-only}}, where the upsampling kernels are generated from encoder features; 
\mbox{3) \textit{encoder-decoder}}, where the upsampling kernels are generated from the concatenation of encoder and NN-interpolated decoder features. 
We report Mask IoU (mIoU)~\cite{everingham2010pascal} and Boundary IoU (bIoU)~\cite{cheng2021boundary} for segmentation, and report Peak Signal-to-Noise Ratio (PSNR), Structural SIMilarity index (SSIM), Mean Absolute Error (MAE), and root Mean Square Error (MSE) for reconstruction.
From Table~\ref{tab:seg_and_rec}, one can observe that the encoder-only baseline outperforms the decoder-only one in image reconstruction, but in semantic segmentation the trend is on the contrary. To understand why, we visualize the segmentation masks and reconstructed results in Fig.~\ref{fig:sunrgbd_visual}. We find that in segmentation the decoder-only model tends to produce region-continuous output, while the encoder-only one generates clear mask boundaries but blocky regions; in reconstruction, by contrast, the decoder-only model almost fails and can only generate low-fidelity reconstructions. It thus can be inferred that, encoder features help to predict details, while decoder features contribute to semantic preservation of regions. Indeed, by considering both encoder and decoder features, the resulting mask seems to integrate the merits of the former two, and the reconstructions are also full of details. Therefore, albeit a simple tweak, FADE significantly benefits from generating upsampling kernels with both encoder and decoder features, as illustrated in Fig.~\ref{fig:concept_map}(c).

\begin{table}[!t]\scriptsize
    \caption{Results of semantic segmentation on SUN RGBD and image reconstruction on Fashion MNIST. Best performance is in \textbf{boldface}.}
    \label{tab:seg_and_rec}
    \centering
    \renewcommand{\arraystretch}{0.8}
    \addtolength{\tabcolsep}{11.5pt}
    \begin{tabular}{@{}lcc|cccc@{}}
    \toprule
    & \multicolumn{2}{c|}{$\tt Segmentation$}  & \multicolumn{4}{c}{$\tt Reconstruction$} \\ 
    & \multicolumn{2}{c|}{$\tt accuracy$ $\tt metric\uparrow$} & \multicolumn{2}{c}{$\footnotesize \tt accuracy$ $\tt metric\uparrow$} & \multicolumn{2}{c}{$\tt error$ $\tt metric\downarrow$}\\
     & mIoU & bIoU & PSNR & SSIM & MAE & MSE\\ 
    \midrule
    decoder-only & 37.00 & 25.61 & 24.35 & 87.19 & 0.0357 & 0.0643 \\
    encoder-only & 36.71 & 27.89 & 32.25 & 97.73 & 0.0157 & 0.0257 \\
    encoder-decoder & \textbf{37.59} & \textbf{28.80} & \textbf{33.83} & \textbf{98.47} & \textbf{0.0122} & \textbf{0.0218} \\ 
    \bottomrule
    \end{tabular}
\end{table}

\paragraph{\textbf{Semi-shift Convolution.}}

\begin{figure}[t]
	\centering
	\includegraphics[width=\linewidth]{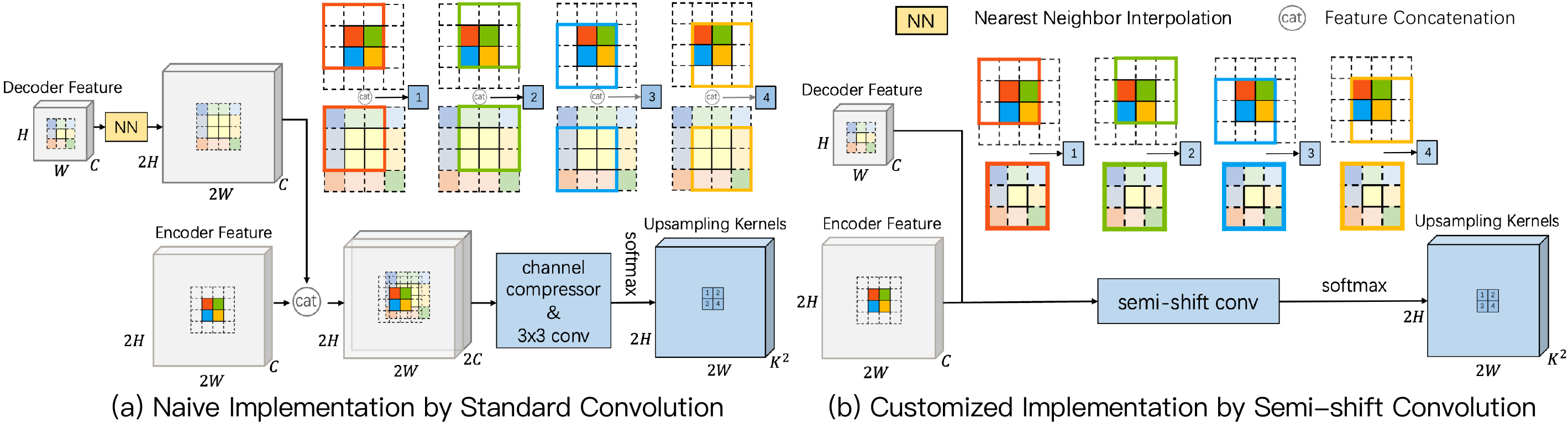}
	\caption{\textbf{Two forms of implementations for generating upsampling kernels}. Naive implementation requires matching resolution with explicit feature interpolation and concatenation, followed by channel compression and standard convolution for kernel prediction. Our customized implementation simplifies the whole process with only semi-shift convolution.}
	\label{fig:semi-shift_conv}
\end{figure}

Given encoder and decoder features, we next address how to use them to generate upsampling kernels. We investigate two implementations: a naive implementation and a customized implementation. The key difference between them is how each decoder feature point spatially corresponds to each encoder feature point. The naive implementation shown in Fig.~\ref{fig:semi-shift_conv}(a) includes four operations: i) feature interpolation, ii) concatenation, iii) channel compression, iv) standard convolution for kernel generation, and v) $\softmax$ normalization. As aforementioned in Section \ref{sec:insights}, naive interpolation can have a few problems. To address them, we present semi-shift convolution that simplifies the first four operations above into a unified operator, which is schematically illustrated in Fig.~\ref{fig:semi-shift_conv}(b). Note that the $4$ convolution windows in encoder features all correspond to the same window in decoder features. This design has the following advantages: 
1) the role of control in the kernel generation is made clear where the control of the variance of $2\times2$ neighbors is moved to encoder features completely;
2) the receptive field of decoder features is kept consistent with that of encoder features; 
3) memory cost is reduced, because semi-shift convolution directly operates on low-resolution decoder features, without feature interpolation;
4) channel compression and $3\times3$ convolution can be merged in semi-shift convolution.
Mathematically, the single window processing with naive implementation or semi-shift convolution has an identical form if ignoring the content of feature maps.
For example, considering the top-left window (`1' in Fig.~\ref{fig:semi-shift_conv}), the (unnormalized) upsampling kernel weight has the form
\begin{align}\footnotesize
    \label{eq:1}
  w_m &= \sum\limits_{l=1}^{d}\sum\limits_{i=1}^{h}\sum\limits_{j=1}^{h}\beta_{ijlm}\left(\sum\limits_{k=1}^{2C}\alpha_{kl}x_{ijk} + a_l\right) + b_m\\
  \label{eq:2}
  &= \sum\limits_{l=1}^{d}\sum\limits_{i=1}^{h}\sum\limits_{j=1}^{h}\beta_{ijlm}\left(\sum\limits_{k=1}^{C}\alpha_{kl}^{\tt en}x_{ijk}^{\tt en} + \sum\limits_{k=1}^{C}\alpha_{kl}^{\tt de}x_{ijk}^{\tt de} + a_l\right) + b_m\\
  \label{eq:3}
  &= \sum\limits_{l=1}^{d}\sum\limits_{i=1}^{h}\sum\limits_{j=1}^{h}\beta_{ijlm}\sum\limits_{k=1}^{C}\alpha_{kl}^{\tt en}x_{ijk}^{\tt en} + \sum\limits_{l=1}^{d}\sum\limits_{i=1}^{h}\sum\limits_{j=1}^{h}\beta_{ijlm}\left(\sum\limits_{k=1}^{C}\alpha_{kl}^{\tt de}x_{ijk}^{\tt de} + a_l\right) + b_m
\end{align}
where $w_m, m=1,...,K^2$, is the weight of the upsampling kernel, $K$ the upsampling kernel size, $h$ the convolution window size, $C$ the number of input channel dimension of encoder and decoder features, and $d$ the number of compressed channel dimension. $\alpha_{kl}^{\tt en}$ and $\{\alpha_{kl}^{\tt de}, a_l\}$ are the parameters of $1\times1$ convolution specific to encoder and decoder features, respectively, and $\{\beta_{ijlm}, b_m\}$ the parameters of $3\times3$ convolution. Following CARAFE, we fix $h=3$, $K=5$ and $d = 64$.
 
According to Eq.~\eqref{eq:3}, by the linearity of convolution, Eq.~\eqref{eq:1} and Eq.~\eqref{eq:2} are equivalent to applying two distinct $1\times1$ convolutions to $C$-channel encoder and $C$-channel decoder features, respectively, followed by a shared $3\times3$ convolution and summation. Eq.~\eqref{eq:3} allows us to process encoder and decoder features without matching their resolution. To process the whole feature map, the window can move $s$ steps on encoder features but only $\lfloor s/2 \rfloor$ steps on decoder features. This is why the operator is given the name `semi-shift convolution'. To implement this efficiently, we split the process to $4$ sub-processes; each sub-process focuses on the top-left, top-right, bottom-left, and bottom-right windows, respectively. Different sub-processes have also different prepossessing strategies. For example, for the top-left sub-process, we add full padding to the decoder feature, but only add padding on top and left to the encoder feature. Then all the top-left window correspondences can be satisfied by setting stride of $1$ for the decoder feature and $2$ for the encoder feature. Finally, after a few memory operations, the four sub-outputs can be reassembled to the expected upsampling kernel, and the kernel is used to reassemble decoder features to generate pre-upsampled features, as shown in Fig.~\ref{fig:pipeline}(b).
 
\begin{figure*}[!t]
\centering
\makeatletter\def\@captype{table}\makeatother
\begin{minipage}[!t]{0.4\textwidth}
\scriptsize
    \caption{The results on the Weizmann Horse dataset.}
    \centering
    \renewcommand{\arraystretch}{0.9} 
    \addtolength{\tabcolsep}{10pt}
    \begin{tabular}{@{}lc@{}}
        \toprule
        SegNet -- baseline & mIoU\\
        \midrule
        Unpooling & 93.42\\
        IndexNet~\cite{lu2019indices} & 93.00\\
        NN & 89.15\\ 
        CARAFE~\cite{jiaqi2019carafe} & 89.29\\
        NN + Gate & 95.26\\
        CARAFE + Gate & 95.25\\
        \bottomrule
    \end{tabular}
    \label{tab:horse}
\end{minipage}
\quad
\makeatletter\def\@captype{figure}\makeatother
\begin{minipage}[!t]{0.5\textwidth}
    \centering
    \includegraphics[width=1\linewidth]{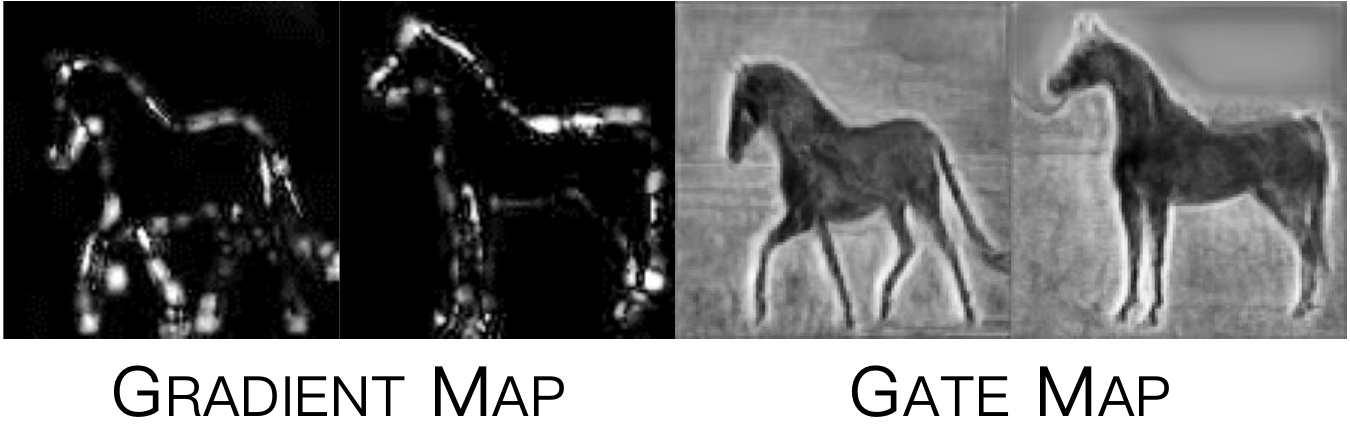}
    \caption{Gradient maps and gate maps of horses.}
    \label{fig:horse}
\end{minipage}
\end{figure*}

\paragraph{\textbf{Extracting Fine Details from Encoder Features.}}
Here we further introduce a gating mechanism to complement fine details from encoder features to pre-upsampled features. We again use some experimental observations to showcase our motivation. We use a binary image segmentation dataset, Weizmann Horse~\cite{borenstein2002class}. The reasons for choosing this dataset are two-fold: 
(1) visualization is made simple; 
(2) the task is simple such that the impact of feature representation can be neglected. 
When all baselines have nearly perfect region predictions, the difference in detail prediction can be amplified. We use SegNet pretrained on ImageNet as the baseline and alter only the upsampling operators.
Results are listed in Table \ref{tab:horse}. 
An interesting phenomenon is that CARAFE works almost the same as NN interpolation and even falls behind the default unpooling and IndexNet. An explanation is that the dataset is too simple such that the region smoothing property of CARAFE is wasted, but recovering details matters. 

A common sense in segmentation is that, the interior of a certain class would be learned fast, while mask boundaries are difficult to predict.  
This can be observed from the gradient maps w.r.t.\ an intermediate decoder layer, as shown in Fig.~\ref{fig:horse}. During the middle stage of training, most responses are near boundaries. 
Now that gradients reveal the demand of detail information, feature maps would also manifest this requisite with some distributions, e.g., in multi-class semantic segmentation a confident class prediction in a region would be a unimodal distribution along the channel dimension, and an uncertain prediction around boundaries would likely be a bimodal distribution. Hence, we assume that all decoder layers have gradient-imposed distribution priors and can be encoded to inform the requisite of detail or semantic information. In this way fine details can be chosen from encoder features without hurting the semantic property of decoder features. Hence, instead of directly skipping encoder features as in feature pyramid networks~\cite{lin2017feature}, we introduce a gating mechanism~\cite{cho2014properties} to selectively refine pre-upsampled features using encoder features, conditioned on decoder features. The gate is generated through a $1\times1$ convolution layer, a NN interpolation layer, and a $\tt sigmoid$ function. As shown in Fig.~\ref{fig:pipeline}(c), the decoder feature first goes through the gate generator, and the generator then outputs a gate map instantiated in Fig.~\ref{fig:horse}. Finally, the gate map $\bm G$ modulates the encoder feature $\mathcal F_{\tt encoder}$ and the pre-upsampled feature $\mathcal F_{\tt pre-upsampled}$ to generate the final upsampled feature $\mathcal F_{\tt upsampled}$ as
\begin{align}
  \mathcal F_{\tt upsampled} = \mathcal F_{\tt encoder} \cdot \bm G + \mathcal F_{\tt pre-upsampled} \cdot (1-\bm G)\,.
\end{align}
From Table~\ref{tab:horse}, the gating mechanism works on both NN and CARAFE.

\section{Results and Discussions}

Here we formally validate FADE on large-scale dense prediction tasks, including image matting 
and semantic segmentation. 
We also conduct ablation studies to justify each design choice of FADE. In addition, we analyze computational complexity in terms of parameter counts and GFLOPs.

\subsection{Image Matting}
Image matting~\cite{xu2017deep} is chosen as the representative of the detail-sensitive task. It requires a model to estimate the accurate alpha matte that smoothly splits foreground from background. Since ground-truth alpha mattes can exhibit significant differences among local regions, estimations are sensitive to a specific upsampling operator used~\cite{lu2019indices,dai2021learning}.

\paragraph{\textbf{Data Set, Metrics, Baseline, and Protocols.}} 
We conduct experiments on the Adobe Image Matting dataset~\cite{xu2017deep}, whose training set has $431$ unique foreground objects and ground-truth alpha mattes. Following~\cite{dai2021learning}, instead of compositing each foreground with fixed $100$ background images chosen from MS COCO~\cite{lin2014microsoft}, we randomly choose background images in each iteration and generate composited images on-the-fly. The Composition-1K testing set has $50$ unique foreground objects, and each is composited with $20$ background images from PASCAL VOC~\cite{everingham2010pascal}. We report the widely used Sum of Absolute Differences (SAD), Mean Squared Error (MSE), Gradient (Grad), and Connectivity (Conn) and evaluate them using the code provided by~\cite{xu2017deep}.

A2U Matting~\cite{dai2021learning} is adopted as the baseline. Following~\cite{dai2021learning}, the baseline network adopts a backbone of the first $11$ layers of ResNet-34 with in-place activated batchnorm~\cite{bulo2018place} and a decoder consisting of a few upsampling stages with shortcut connections. Readers can refer to~\cite{dai2021learning} for the detailed architecture. We use max-pooling at downsampling stages and replace upsampling operators with FADE. We strictly follow the training configurations and data augmentation strategies used in~\cite{dai2021learning}.



\begin{table}[t] \scriptsize
    \caption{Image matting and semantic segmentation results on the Adobe Composition-1k and ADE20K data sets. $\Delta$Param. indicates the additional number of parameters compared with the bilinear baseline. Best performance is in \textbf{boldface}.}
    \centering
    \renewcommand{\arraystretch}{0.9}
    \addtolength{\tabcolsep}{5pt}
    \begin{tabular}{@{}lccccl|ccl@{}}
    \toprule
        A2U Matting/ & \multicolumn{5}{c|}{$\tt Matting$ -- $\tt error\downarrow$} & \multicolumn{3}{c}{$\tt Segm$ -- $\tt accuracy\uparrow$}  \\
        SegFormer & SAD & MSE & Grad & Conn &  $\Delta$Param. & mIoU & bIoU & $\Delta$Param.\\
        \midrule
        Bilinear & 37.31 & 0.0103 & 21.38 & 35.39 & 8.05M & 41.68 & 27.80 & 13.7M\\
        CARAFE~\cite{jiaqi2019carafe} & 41.01 & 0.0118 & 21.39 & 39.01 & +0.26M & 42.82 & 29.84 & +0.44M\\
        IndexNet~\cite{lu2019indices} & 33.36 & 0.0086 & 16.17 & 30.62 & +12.26M & 41.50 & 28.27 & +12.60M\\
        A2U~\cite{dai2021learning} & 32.05 & 0.0081 & 15.49 & 29.21 & +38K & 41.45 & 27.31 & +0.12M\\
        FADE (Ours) & \textbf{31.10} & \textbf{0.0073} & \textbf{14.52} & \textbf{28.11} & +0.12M & \textbf{44.41} & \textbf{32.65} & +0.29M\\
    \bottomrule
    \end{tabular}
    \label{tab:performance}
\end{table}

\paragraph{\textbf{Matting Results.}} We compare FADE with other state-of-the-art upsampling operators. Quantitative results are shown in Table \ref{tab:performance}. Results show that FADE consistently outperforms other competitors in all metrics, with also few additional parameters. It is worth noting that IndexNet and A2U are strong baselines that are delicately designed upsampling operators for image matting. Also the worst performance of CARAFE indicates that upsampling with only decoder features cannot meet a detail-sensitive task. Compared with standard bilinear upsampling, FADE invites $16\%\sim32\%$ relative improvement, which suggests upsampling can indeed make a difference, and our community should shift more attention to upsampling. Qualitative results are shown in Fig.~\ref{fig:seg_matte_intro}. FADE generates a high-fidelity alpha matte.

\subsection{Semantic Segmentation}
Semantic segmentation is chosen as the representative region-sensitive task. To prove that FADE is architecture-independent, SegFormer~\cite{xie2021segformer}, a recent transformer based segmentation model, is used as the baseline.

\paragraph{\textbf{Data Set, Metrics, Baseline, and Protocols.}} 
We use the ADE20K dataset \cite{zhou2017scene}, which is a standard benchmark used to evaluate segmentation models. ADE20K covers $150$ fine-grained semantic concepts, including $20210$ images in the training set and $2000$ images in the validation set. In addition to reporting the standard Mask IoU (mIoU) metric~\cite{everingham2010pascal}, we also include the Boundary IoU (bIoU) metric~\cite{cheng2021boundary} to assess boundary quality.

SegFormer-B1~\cite{xie2021segformer} is chosen by considering both the effectiveness and computational sources at hand. We keep the default model architecture in SegFomer except for modifying the upsampling stage in the MLP head. All training settings and implementation details are kept the same as in~\cite{xie2021segformer}.

\paragraph{\textbf{Segmentation Results.}}
Quantitative results of different upsampling operators are also listed in Table~\ref{tab:performance}. Similar to matting, FADE is the best performing upsampling operator in both mIoU and bIoU metrics. Note that, among compared upsampling operators, FADE is the only operator that exhibits the task-agnostic property. A2U is the second best operator in matting, but turns out to be the worst one in segmentation. CARAFE is the second best operator in segmentation, but is the worst one in matting. This implies that current dynamic operators still have certain weaknesses to achieve task-agnostic upsampling. Qualitative results are shown in Fig.~\ref{fig:seg_matte_intro}. FADE generates high-quality prediction both within mask regions and near mask boundaries.

\begin{table*}[!t] \scriptsize
    \caption{Ablation study on the source of features, the way for upsampling kernel generation, and the effect of the gating mechanism. Best performance is in \textbf{boldface}. en: encoder; de: decoder.}
    \label{tab:ablation_study}
    \centering
    \renewcommand{\arraystretch}{0.9} 
    \addtolength{\tabcolsep}{2.6pt}
    \begin{tabular}{@{}lccccccc|cc@{}}
        \toprule
        No. & \multicolumn{3}{c}{A2U Matting / SegFormer} & \multicolumn{4}{c|}{$\tt Matting$ - $\tt error\downarrow$} & \multicolumn{2}{c}{$\tt Segm$ - $\tt accuracy\uparrow$}\\
        & source of feat. & kernel gen. & fusion & SAD & MSE & Grad & Conn & mIoU & bIoU\\
        \midrule
        B1 & en &  & & 34.22 & 0.0087 & 15.90 & 32.03 & 42.75 & 31.00\\
        B2 & de &  & & 41.01 & 0.0118 & 21.39 & 39.01 & 42.82 & 29.84\\
        B3 & en \& de & naive & & 32.41 & 0.0083 & 16.56 & 29.82 & 43.27 & 31.55\\
        B4 & en \& de & semi-shift & & 31.78 & 0.0075 & 15.12 & 28.95 & 43.33 & 32.06\\
        B5 & en \& de & semi-shift & skipping & 32.64 & 0.0076 & 15.90 & 29.92 & 43.22 & 31.85\\
        B6 & en \& de & semi-shift & gating & \textbf{31.10} & \textbf{0.0073} & \textbf{14.52} & \textbf{28.11} & \textbf{44.41} & \textbf{32.65}\\
        \bottomrule
    \end{tabular}
\end{table*}

\subsection{Ablation Study}

Here we justify how performance is affected by the source of features, the way for upsampling kernel generation, and the use of the gating mechanism. We build six baselines based on FADE:
\begin{enumerate}
    \item[1)] B1: \textit{encoder-only}. Only encoder features go through $1\times1$ convolution for channel compression ($64$ channels), followed by $3\times3$ convolution layer for kernel generation;
    
    \item[2)] B2: \textit{decoder-only}. This is the CARAFE baseline~\cite{jiaqi2019carafe}. Only decoder features go through the same $1\times1$ and $3\times3$ convolution for kernel generation, followed by Pixel Shuffle as in CARAFE due to different spatial resolution;
    
    \item[3)] B3: \textit{encoder-decoder-naive}. NN-interpolated decoder features are first concatenated with encoder features, and then the same two convolutional layers are applied;
    
    \item[4)] B4: \textit{encoder-decoder-semi-shift}. Instead of using NN interpolation and standard convolutional layers, we use semi-shift convolution to generate kernels directly as in FADE;
    
    \item[5)] B5: B4 with \textit{skipping}. We directly skip the encoder features as in feature pyramid networks~\cite{lin2017feature}; 
    
    \item[6)] B6: B4 with \textit{gating}. The full implementation of FADE.
    
\end{enumerate} 
Results are shown in Table \ref{tab:ablation_study}. By comparing B1, B2, and B3, the experimental results give a further verification on the importance of both encoder and decoder features for upsampling kernel generation. By comparing B3 and B4, the results indicate a clear advantage of semi-shift convolution over naive implementation in the way of generating upsampling kernels. As aforementioned, the rationale that explains such a superiority can boil down to the granular control of the contribution of each feature point in kernels (Section \ref{sec:fade}). We also note that, even without gating, the performance of FADE already surpasses other upsampling operators (B4 vs.\ Table~\ref{tab:performance}), which means the task-agnostic property is mainly due to the joint use of encoder and decoder features and the semi-shift convolution. In addition, skipping is clearly not the optimal way to move encoder details to decoder features, at least worse than the gating mechanism (B5 vs.\ B6).

\begin{figure}[!t]
\centering
\captionsetup{singlelinecheck=true}
     \begin{subfigure}[b]{0.32\textwidth}
         \centering
         \includegraphics[width=\textwidth]{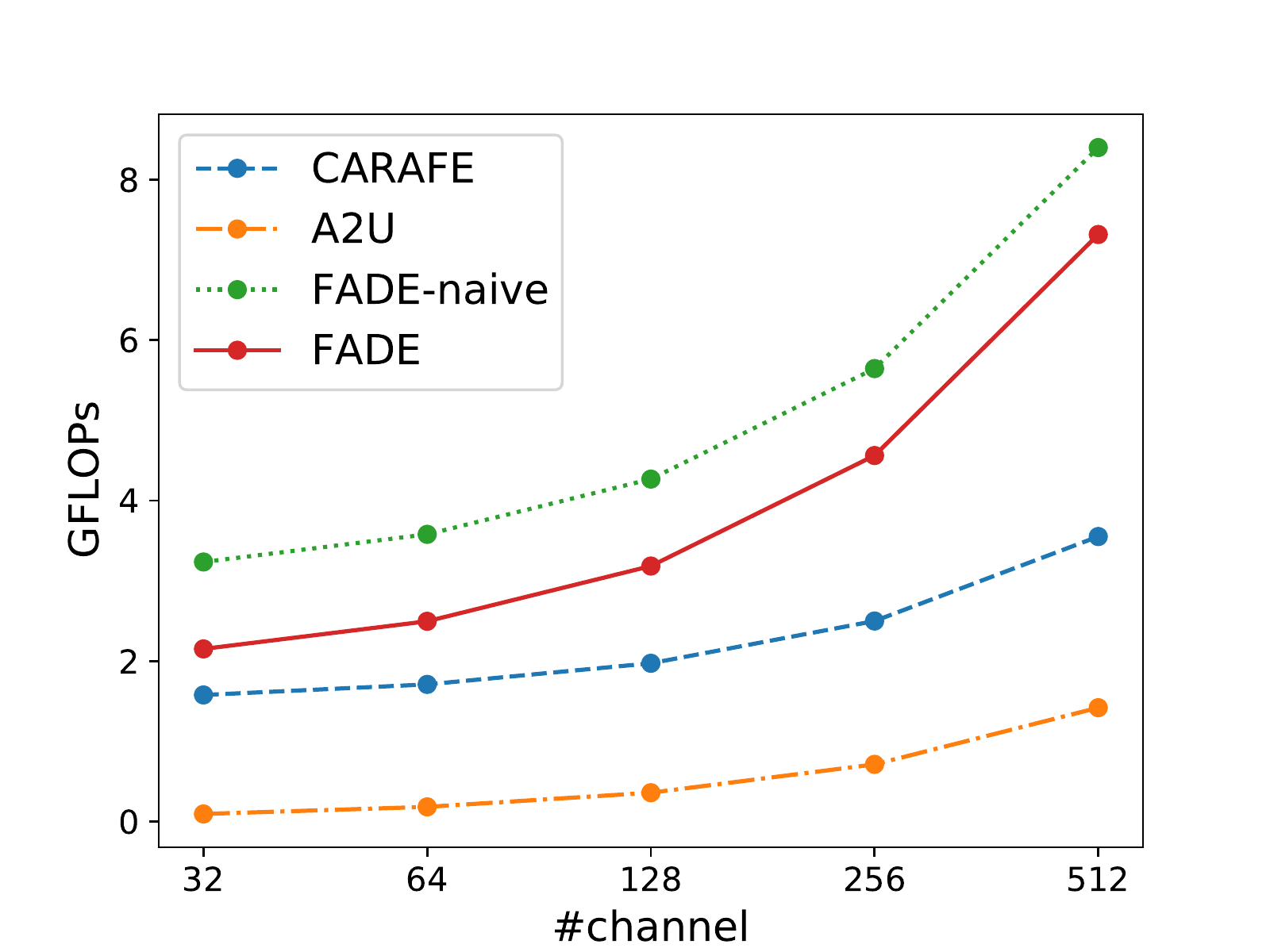}
         \caption{resolution=$112\times112$}
         \label{fig:y equals x}
     \end{subfigure}
     \hfill
     \begin{subfigure}[b]{0.32\textwidth}
         \centering
         \includegraphics[width=\textwidth]{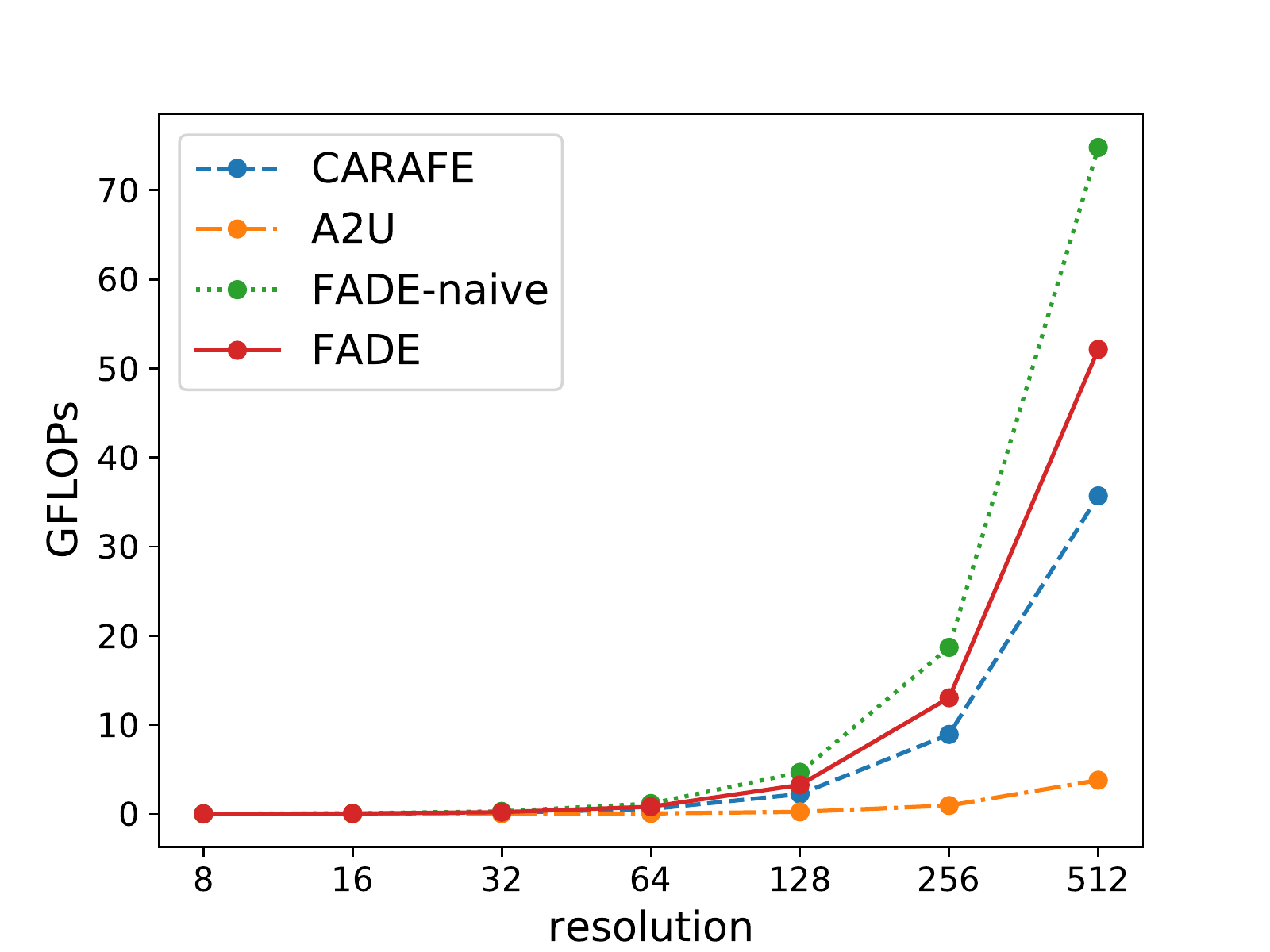}
         \caption{channel=$64$}
     \end{subfigure}
     \hfill
     \begin{subfigure}[b]{0.325\textwidth}
         \centering
         \includegraphics[width=\textwidth]{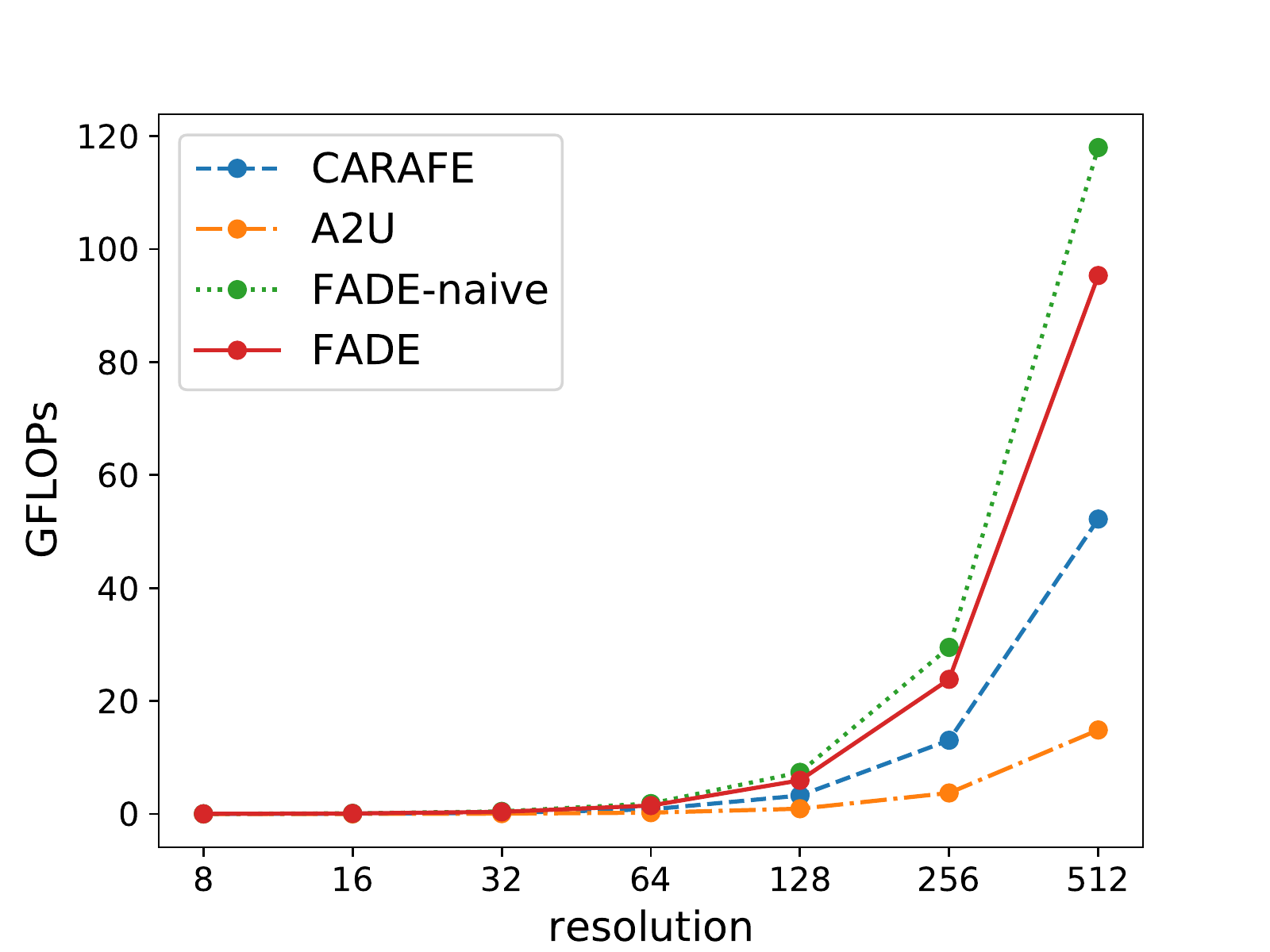}
         \caption{channel=$256$}
         \label{fig:five over x}
     \end{subfigure}
    \caption{GFLOPs comparison between FADE and other upsampling operators.}
    \label{fig:gflops}
\end{figure}

\subsection{Comparison of Computational Overhead}
A favorable upsampling operator, being part of overall network architecture, should not significantly increase the computation cost. This issue is not well addressed in IndexNet as it significantly increases the number of parameters and computational overhead~\cite{lu2019indices}. Here we measure GFLOPs of some upsampling operators by i) changing number of channels given fixed spatial resolution and by ii) varying spatial resolution given fixed number of channels. Fig.~\ref{fig:gflops} suggests FADE is also competitive in GFLOPs, especially when upsampling with relatively low spatial resolution and low channel numbers. In addition, semi-shift convolution can be considered a perfect replacement of the standard `interpolation+convolution' paradigm for upsampling, not only superior in effectiveness but also in efficiency.


\section{Conclusions}

In this paper, we propose FADE, a novel, plug-and-play, and task-agnostic upsampling operator. For the first time, FADE demonstrates the feasibility of task-agnostic feature upsampling in both region- and detail-sensitive dense prediction tasks, outperforming the best upsampling operator A2U on image matting and the best operator CARAFE on semantic segmentation. With step-to-step analyses, we also share our view points from considering what makes for generic feature upsampling.

For future work, we plan to validate FADE on additional dense prediction tasks and also explore the peer-to-peer downsampling stage. So far, FADE is designed to maintain the simplicity by only implementing linear upsampling, which leaves much room for further improvement, \textit{e.g.}, with additional nonlinearity. In addition, we believe how to strengthen the coupling between encoder and decoder features to enable better cooperation can make a difference for feature upsampling.

\paragraph{\textbf{Acknowledgement}.} This work is supported by the Natural Science Foundation of China under Grant No.\  62106080.

\clearpage
%
%
\bibliographystyle{splncs04}
\bibliography{egbib}

\clearpage

\appendix

\section*{Appendix}

We provide the following contents in this appendix:
\begin{itemize}
    \item[-] Visualization of upsampled features between FADE and CARAFE;
    \item[-] Implementation details of segmentation on the Weizmann Horse data set;
    \item[-] Illustration on how FADE is incorporated into SegFormer;
    \item[-] Additional visualizations of semantic segmentation on ADE20K and image matting on Adobe Composition-1K.
\end{itemize}

\section{Visualization of Upsampled Features}
\begin{figure}[!t]
	\centering
	\includegraphics[width=\linewidth]{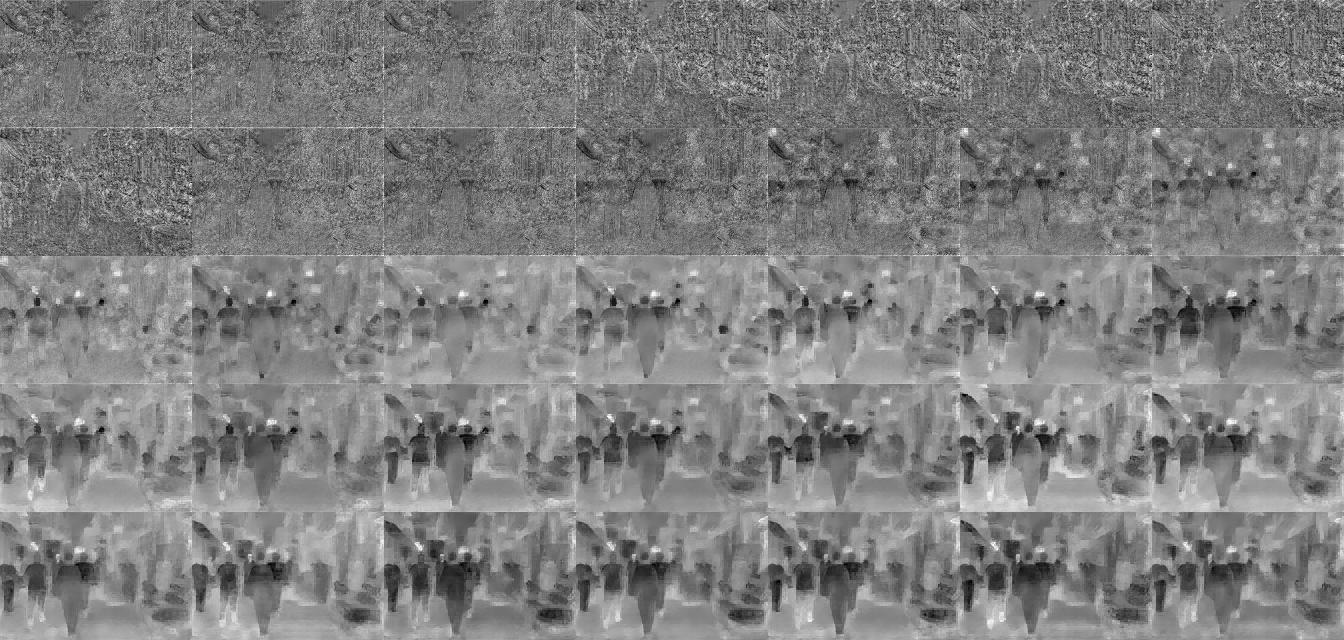}
	\caption{\textbf{Feature maps upsampled by FADE with increased training iterations.} From left to right, from top to bottom, FADE first learns to delineate the outlines of objects and then gradually fills the interior regions.}
	\label{fig:fade}
	\vspace{20pt}
	\includegraphics[width=\linewidth]{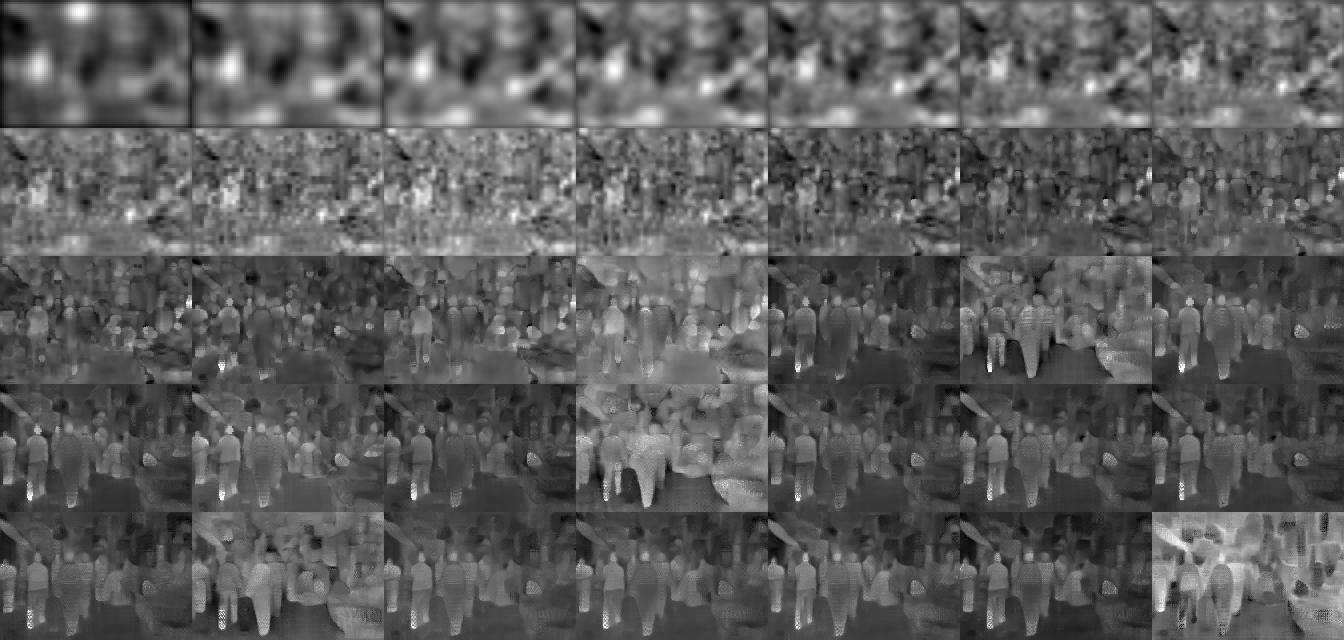}
	\caption{\textbf{Feature maps upsampled by CARAFE with increased training iterations.} From left to right, from top to bottom, CARAFE starts from the interior and then spreads outside.}
	\label{fig:carafe}
\end{figure}

We visualize the upsampled feature maps w.r.t.\ CARAFE and FADE in SegFormer. We select one checkpoint for every $100$ iterations 
in the range from the $100$-th to $3000$-th iteration. We also highlight the $320$-th, $330$-th, $340$-th, $350$-th, and $360$-th iteration, because we observe fast variances of the feature maps during this period. We 
compute the average response along the channel dimension 
and normalize it to $[0,255]$.
From Figures~\ref{fig:carafe} and~\ref{fig:fade}, we can see that the two upsampling operators have different behaviors: FADE first learns to delineate the outlines of objects and then gradually fills the interior regions, while CARAFE focuses on the interior initially and then spreads outside slowly. We think the reason is that the gating mechanism is relatively simple and learns fast. By the way, one can see that there is `checkerboard artifacts' in the visualizations of CARAFE due to the adoption of Pixel Shuffle.

\section{Implementation Details for Segmentation on the Weizmann Horse Data Set}
There are $328$ images with ground truth masks in the Weizmann Horse data set, in which we randomly choose $85\%$ images as the training set, and the rest as the testing set. SegNet pretrained on ImageNet is used as the basic architecture and we only modify its upsampling operator during experiments. Images are resized to $224\times224$. We use the cross entropy loss. The batch size is set to $4$. We use SGD with a momentum of 0.9 as the optimizer. We set the initial learning rate as $0.01$ and decay the rate at the $35$-th and $45$-th epoch to $0.001$ and $0.0001$, respectively.

For visualization, we output the gradient and the feature maps from an intermediate decoding layer (NOT the last layer). For the gradient maps, considering that there exist positive or negative values among different regions, we pass them through a ReLU function first and then sum all the channels. For gated feature maps, we select some representative ones from the channels. Both maps are normalized to $[0,255]$ for visualization.

\section{How FADE Is Incorporated into SegFormer}
\begin{figure}[!t]
	\centering
	\includegraphics[width=\linewidth]{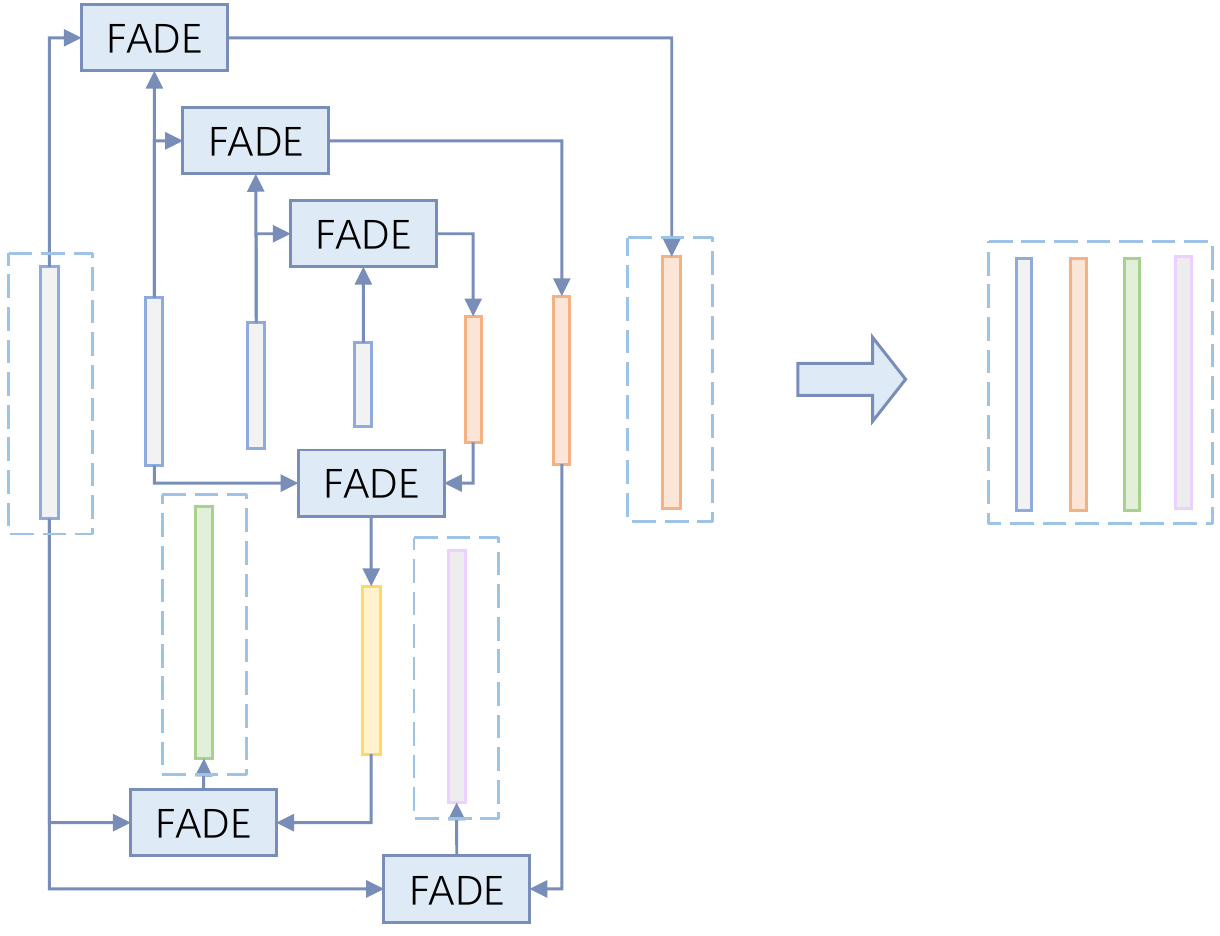}\vspace{-5pt}
	\caption{\textbf{Illustration on how FADE is incorporated into SegFormer.}}
	\label{fig:fade_insert}
\end{figure}

In image matting, the input and output are the same as A2U. In semantic segmentation, as shown in Fig.~\ref{fig:fade_insert}, feature maps of each scale need to be upsampled to $1/4$ of the original image. Therefore, there are $3+2+1=6$ upsampling operators involved in all.

\section{Additional Visualizations}
\begin{figure}[!t]
	\centering
	\includegraphics[width=\linewidth]{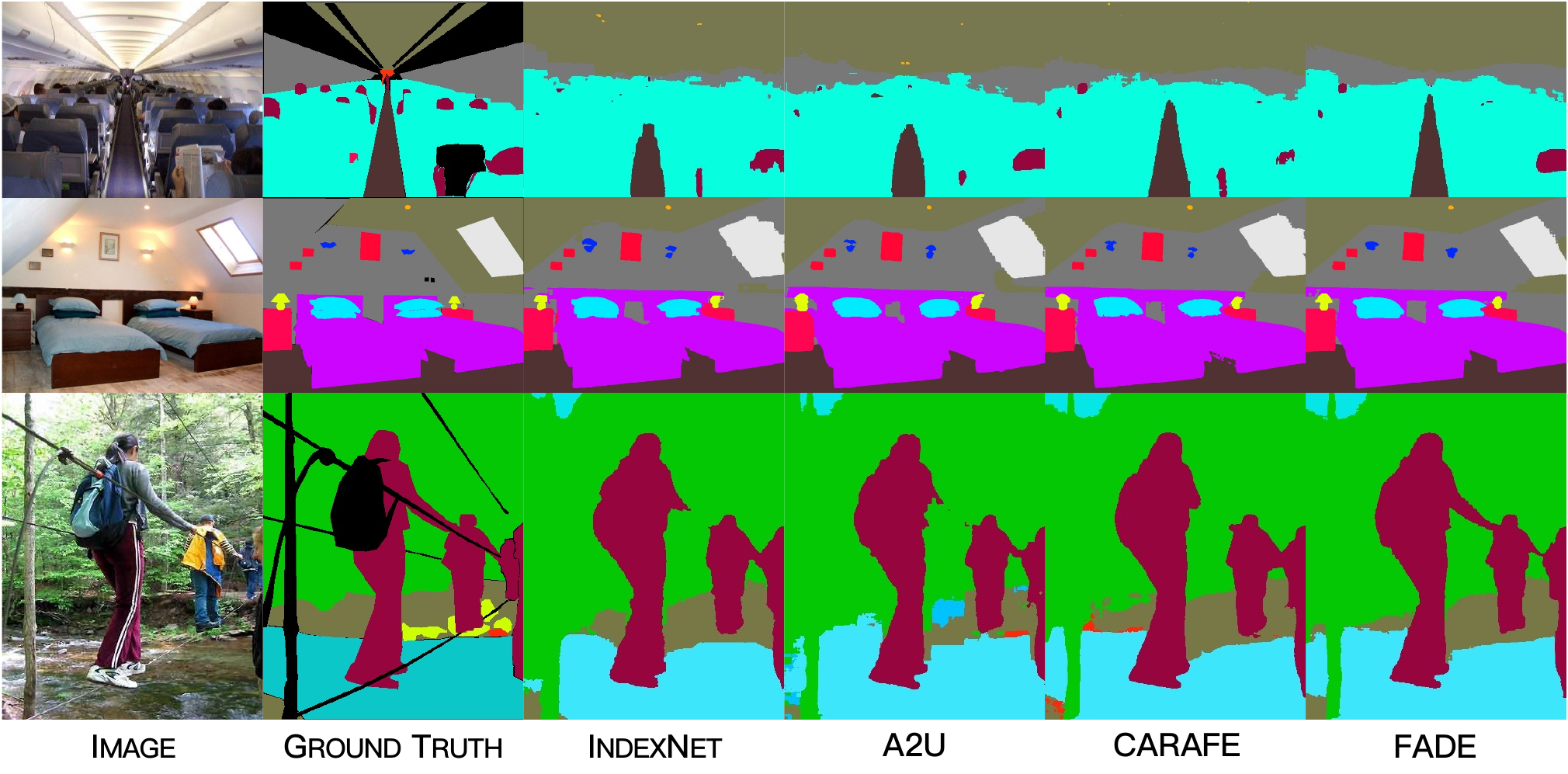}
	\caption{\textbf{Additional visualizations of different upsampling operators on the ADE20K data set.} Compared with other upsampling operators, FADE maintains both regional continuity and boundary accuracy.}
	\label{fig:ade20k_visual}
\end{figure}

\begin{figure}[!t]
	\centering
	\includegraphics[width=\linewidth]{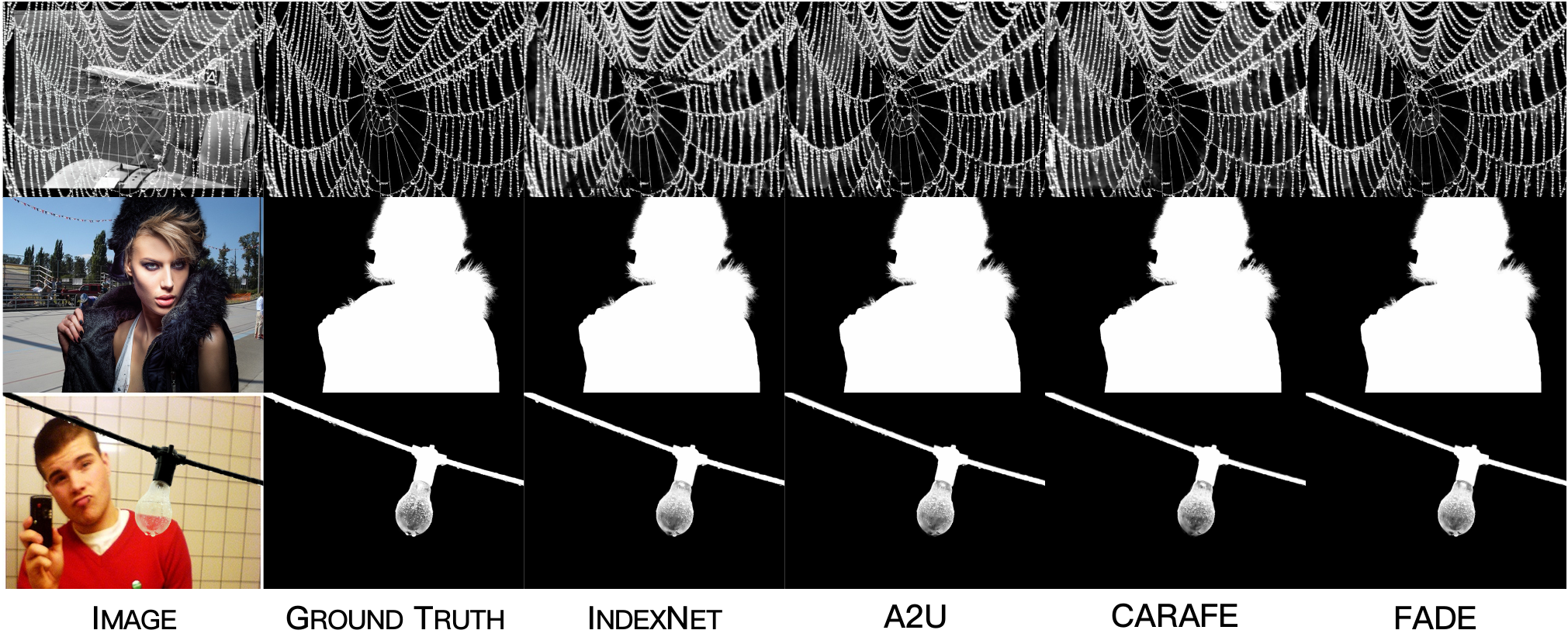}
	\caption{\textbf{Additional visualizations on the Adobe Composition-1K testing set.} Compared with other upsampling operators, FADE invites better detail delineation, \textit{e.g.}, the water drop below the bulb.}
	\label{fig:adobe_visual}
\end{figure}

Here we give additional visualizations on the ADE20K (Fig.~\ref{fig:ade20k_visual}) and the Adobe Composition 1K (Fig.~\ref{fig:adobe_visual}) data sets. In segmentation, `SegFormer+FADE' exhibits not only improved regional integrity but also sharp and consistent edges. In matting, FADE also contributes significantly to detail recovery, \textit{e.g.}, the water drop below the bulb.

\end{document}